\documentclass[10pt]{amsart}
\usepackage{kept}
\usepackage{times}
\usepackage{latexsym}
\usepackage{graphicx}

\begin{document}
\title[Title]{ A chain  dictionary method  for Word Sense Disambiguation and applications.}

\author{ Doina T\u atar$^{(1)}$}
\address{$^{1}$ University "Babes-Bolyai", Cluj-Napoca}
\email{dtatar@cs.ubbcluj.ro}

\author{ Gabriela \c Serban$^{(2)}$}
\address{$^{2}$ University "Babes-Bolyai", Cluj-Napoca}
\email{gabis@cs.ubbcluj.ro}

\author{Andreea Mihi\c s$^{(3)}$}
\address{$^{3}$ University "Babes-Bolyai", Cluj-Napoca}
\email{mihis@cs.ubbcluj.ro}

\author{Mihaiela Lupea$^{(4)}$}
\address{$^{4}$ University "Babes-Bolyai", Cluj-Napoca}
\email{mlupea@cs.ubbcluj.ro}

\author{Dana Lup\c sa$^{(5)}$}
\address{$^{5}$ University "Babes-Bolyai", Cluj-Napoca}
\email{dlupsa@cs.ubbcluj.ro}

\author{Militon Fren\c tiu$^{(6)}$}
\address{$^{6}$ University "Babes-Bolyai", Cluj-Napoca}
\email{mfrentiu@cs.ubbcluj.ro}


\date{}

\begin{abstract}

  A large class of unsupervised algorithms for Word Sense Disambiguation (WSD)
  is that of dictionary-based methods. Various algorithms have as the root
  Lesk's algorithm, which exploits the sense definitions in the dictionary directly.
  Our approach uses the lexical base  WordNet  for a new algorithm originated
  in Lesk's, namely {\it chain algorithm for disambiguation} of all words (CHAD). We show
   how translation from a language into another one and also
   text entailment verification could be accomplished by this disambiguation.

\end{abstract}


 \maketitle

\section{The polysemy}

Word sense disambiguation  is the process of
 identifying the correct sense of words in particular contexts.
 The solving of WSD seems to be  AI complete
 ( that means its  solution requires a solution to all the general
 AI problems of representing and reasoning about arbitrary) and it
 is one of the most important open problems in NLP \cite{nancy},\cite{jm},\cite{Kil},
 \cite{Resnik},\cite{senseval},\cite{ts}.
 In the electronical on-line dictionary WordNet, the most well-developed
and widely used lexical database for English, the polysemy of
different category of words is presented in order as: the highest
for verbs, then for nouns, and the lowest for adjectives and
adverbs.
  Usually, the process of disambiguation is realized for a single, target word.
 One would expect the words closest to the target word to be of greater semantical
   importance for it than the other words in the text. The context  is
   hence a source of information  to identify
   the meaning of the polysemous words.
   The  contexts may be used in two  ways: a) as {\it bag of words}, without consideration
   of
    relationships with  the target word in terms of distance, grammatical relations, etc.;
    b) with relational information.
    The {\it bag of words} approach works better for nouns than verbs but is
    less effective than methods that take other relations in consideration.
        Studies about syntactic relations determined some interesting
     conclusions: verbs derive more disambiguation information from their objects than from their
     subjects, adjectives derive almost all disambiguation information from
     the nouns they modify, and nouns are best
     disambiguated by directly adjacent adjectives or nouns \cite{nancy}.
       All these advocate that a global approach (disambiguation of  all words)
       helps to disambiguate each POS.

      In this paper we propose a global disambiguation  algorithm called {\bf chain algorithm}
      for disambiguation, CHAD,
      which presents elements of both
      points of view about a context: because this algorithm is $ order \,\, sensitive$
       it belongs to the  class of algorithms which depend of  relational
       information;
       in the same time it doesn't require syntactic analysis and syntactic parsing.

       In section 2 of this paper we review Lesk's algorithm for WSD. In section 3 we present "triplet" algorithm for three
       words and CHAD algorithm. In section 4 we describe some experiments and evaluations
       with CHAD. Section 5 introduces some conclusions of using the  CHAD for translation
       (here from Romanian language to English) and for text entailment verification.
         Section 6 draws some conclusions and further work.

\section{Dictionary-based methods}
  Work in WSD reached a turning point in the 1980s when large-scale lexical resources,
  such as  machine readable   dictionaries, became widely available.
     One of the best known dictionary-based method is that of
    Lesk (1986). It starts from the idea that a word's dictionary definition is a good
    indicator for the senses of this word and  uses the definition in the dictionary directly.

       Let us remember basic algorithm of Lesk \cite{manshu}:

     Suppose that for a polysemic target word $w$ there are in a dictionary $Ns$ senses

     $s_1,
     s_2, \cdots, s_{Ns}$ given in  an equal number of definitions   $D_1,
     D_2, \cdots, D_{Ns}$. Here we mean by  $D_i$ the set of words contained
     in the $i$-th definition.

     Consider that the new context to be disambiguated is $c_{new}$.
 The  {\bf reduced form} of Lesk's algorithm is:

 \hspace{0.5cm} {\it   for $ k =1,Ns $ do }

\hspace{1cm} $score(s_k)= \mid D_k \cap ( \cup_{v_j \in c_{new}}
\{v_j\} ) \mid $

\hspace{0.5cm} {\it endfor}

\hspace{0.5cm} {\it Calculate $ s'= argmax_k  score (s_k)$}

The score of a sense is the number of words that are shared by the
different sense definitions (glosses) and the context. A target
word is assigned that sense whose gloss shares the largest number
of words.






The algorithm of Lesk was successfully developed in \cite{pedban}
 by using WordNet dictionary for English.
  It was created by hand in 1990s and includes definitions
 (glosses) for individual senses of words, as in a dictionary.
 Additionally it defines groups of synonymous words representing the same lexical concept (synset)
 and organizes them into a conceptual hierarchy.
   The paper \cite{pedban} uses this conceptual hierarchy for improving the original
   Lesk's method by augmenting the definitions with non-gloss information: synonyms, examples
 and glosses of related words (hypernyms, hyponyms). Also, the authors introduced
 a novel overlap measure between glosses which favorites multi-word matching.

\section{Chain algorithm for word sense disambiguation - CHAD.}

      First of all we present an algorithm for disambiguation of a triplet. In a sense, our
      triplet algorithm is similar with global disambiguation algorithm for a window of
      two words around a target word given \cite{pedban}. Instead, our CHAD realizes
       disambiguation of all-words in a text with any length,  ignoring the notion of "window" and "target word"
        and target word  in similar studies, all that without increasing the computational complexity.

The algorithm for disambiguation of a triplet of words $w_1 w_2
w_3$ for Dice measure is the following:

\begin{small}


 \hspace{0.5cm}       begin

\hspace{1cm}   for each   sense $s_{w_1}^i$  do

 \hspace{1.2cm} for each   sense $s_{w_2}^j$ do

 \hspace{1.4cm}        for each sense $s_{w_3}^k $ do

 \hspace{1.5cm}       $score (i,j,k)=3 \times \frac{\mid D_{w_1} \cap D_{w_2} \cap
  D_{w_3}\mid}
{\mid D_{w_1}\mid + \mid D_{w_2}\mid +   \mid
                                      D_{w_3}\mid}$

\hspace{1.4cm}   endfor

\hspace{1.2cm}       endfor

\hspace{1cm} endfor

\hspace{1cm} $(i^*,j^*,k^*)= argmax_{(i,j,k)}score(i,j,k)$
 \hspace{1cm}           /* sense of $w_1$ is $s_{w_1}^{i^*}$, sense of $w_2$ is
            $s_{w_2}^{j^*}$,
            sense of $w_3$ is $s_{w_3}^{k^*}$ */

\hspace{0.5cm}  end

\end{small}

 For the overlap measure  the score is calculated as:
  $score (i,j,k)=
  \frac{\mid D_{w_1} \cap D_{w_2} \cap  D_{w_3}\mid}
{min (\mid D_{w_1}\mid , \mid D_{w_2}\mid ,   \mid
                                      D_{w_3}\mid)}$
For the Jaccard measure the score is calculates as:
 $score (i,j,k)=
 \frac{\mid D_{w_1} \cap D_{w_2} \cap  D_{w_3}\mid}
{\mid D_{w_1} \cup  D_{w_2} \cup
                                      D_{w_3}\mid}$

 Shortly, CHAD begins  with the disambiguation of a triplet $w_1 w_2 w_3$
 and then adds to the right the following  word to be disambiguated. Hence it disambiguates
 at a time a new triplet, where first two words are already associated with the best senses
 and the disambiguation of the third word depends on these first two words.
 CHAD algorithm for disambiguation of the sentence  $w_1 w_2... w_N$
 is:


\begin{small}
       begin

\hspace{0.3cm} Disambiguate triplet $w_1 w_2 w_3$

\hspace{0.3cm} $i=4$

\hspace{0.3cm} while $i\leq N$ do

\hspace{0.5cm} Calculate $score(s_i)=3 \times \frac{\mid
D_{w_{i-2}}^* \cap D_{w_{i-1}}^* \cap  D_{w_i}^{s_i}\mid} {\mid
D_{w_{i-2}}^*\mid + \mid D_{w_{i-1}}^*\mid +  \mid D_{w_i}^{s_i}
\mid}$

\hspace{0.5cm} Calculate $s_i^* :=argmax_{s_i}score(s_i)$

\hspace{0.5cm} $i:=i+1$

\hspace{0.3cm} endwhile

 end

\end{small}

  Due to the brevity of definitions in WN many values of $\mid D_{w_{i-2}}^* \cap
D_{w_{i-1}}^* \cap  D_{w_i}^{s_i}\mid $ are 0. We attributed the first sense in WN
for $ s_i^*$ in this cases.

 \section{Some experiments with chain algorithm. Experimental evaluation of {\bf CHAD}}

 In this section we shortly describe some experiments that we
 have made in order to validate the proposed chain algorithm
 {\bf CHAD}.

 \subsection{Implementation details}

We have developed an application that implements {\bf CHAD} and
can  be used to:

\begin{itemize}

\item disambiguate words (\ref{res});

\item translate words into Romanian language (\ref{app2});

\item text entailment verification (5.2).

\end{itemize}

The application is written in JDK 1.5.0. and uses \emph{HttpUnit}
1.6.2 API \cite{httpunit}.
 Written in Java, HttpUnit is a free software that emulates the relevant portions
 of browser behavior, including form submission, JavaScript, basic
 http authentication, cookies and automatic page redirection,
 and allows Java test code to examine returned pages either as text,
 an XML DOM, or containers of forms, tables, and links \cite{httpunit}.

We have used \emph{HttpUnit} in order to search WordNet through the dictionary
from \cite{wordnet}. More specifically, the following Java classes
from \cite{httpunit} are used:

\begin{itemize}

\item \emph{WebConversation}. It represents the context for a series
of HTTP requests. This class manages cookies used to maintain session context,
computes relative URLs, and generally emulates the browser behavior needed
to build an automated test of a web site.

\item \emph{WebResponse}. This class represents a response to a
web request from a web server.

\item \emph{WebForm}. This class represents a form in an HTML page.
Using this class we can  examine the parameters defined for the form,
the structure of the form (as a DOM), and the text of the form.
We have used \emph{WebForm} class in order to simulate the submission
of the form with corresponding parameters.

\end{itemize}

\subsection{Results}\label{res}

  We tested our CHAD on 10 files of Brown corpus, which are   POS tagged.
    Recall that WN stores only stems of words. So, we first
    preprocessed  the glosses and the input files, replacing inflected words with
    their stems.

    The reason for  choosing Brown corpus was the possibility offered by SemCor corpus
    (the best known publicly available corpus hand tagged with WN senses)
    to evaluate the results.
  The correct disambiguated words means the disambiguated words as in SemCor.
  We ran separately CHAD for: 1. nouns, 2. verbs, and 3. nouns, verbs, adjectives and adverbs.
  In the case of CHAD addressed to nouns, the output is the sequence  of nouns
  tagged with senses.
  The tag  $noun\#n\#i$ means that for noun $noun$ the WN sense $i$ was found.
  Analogously for
  the case of disambiguation on verbs and of all  POS.
  The results are presented in tables 1 and 2.  As our CHAD algorithm is
  dependent on the length
   of glosses, and as nouns have the longest glosses, the highest precision
    is obtained for nouns.  In Figure 3, the Precision Progress can be traced.
By dropping and rising, the precision finally stabilizes to value
0.767 (for the file  Br-a01). The most interesting part of this
graph is that he shows how this Chain Algorithm works and how the
correct or incorrect disambiguation of first two words from the
first triplet influences the disambiguation of the next words.

   It is known that,  at Senseval 2 contest, only 2 out of the 7 teams (with the
   unsupervised methods) achieved higher precision than the WordNet $1^{st}$
    sense baseline. We compared in figures 1 , 2 and 3 the precision of CHAD
    for 10 files in Brown
    corpus, for Dice, Overlap and Jaccard measures
    with WordNet $1^{st}$ sense.

Comparing the precision obtained with the Overlap Measure and the
precision given by the WordNet $1^{st}$ sense for  10 files of
Brown corpus (Br-a01, Br-a02, Br-11, Br-12, Br-13, Br-14, Br-a15,
Br-b13, Br-b20 and Br-c01), we obtained the following results:

\begin{itemize}
\item  for Nouns, the minimum difference was 0.0077, the maximum
difference was 0.0706, the average difference was 0.0338; \item as
a whole, for 4 files difference was greater or equal to 0.04, and
for 6 files was lower; \item in case of all Parts of Speech, the
minimum difference was 0.0313, the maximum difference was 0.0681,
the average difference was 0.0491; \item   as a whole, for 7 files
difference was greater or equal to 0.04, and for 3 files was
lower; \item   relatively to Verbs, the minimum difference was
0.0078, the maximum difference was 0.0591, the  average difference
was 0.0340;\item  as a whole, for 4 files difference was greater
or equal to 0.04, and for 6 files was lower.
\end{itemize}

Let us remark that in our CHAD the standard concept of windows
better size parameter
    \cite{pedban} is not working: simply, a window is the variable space between the previous
    and the following word in respect to the current word.

\begin{figure}[h]
    \centering
        \includegraphics[scale=0.3]{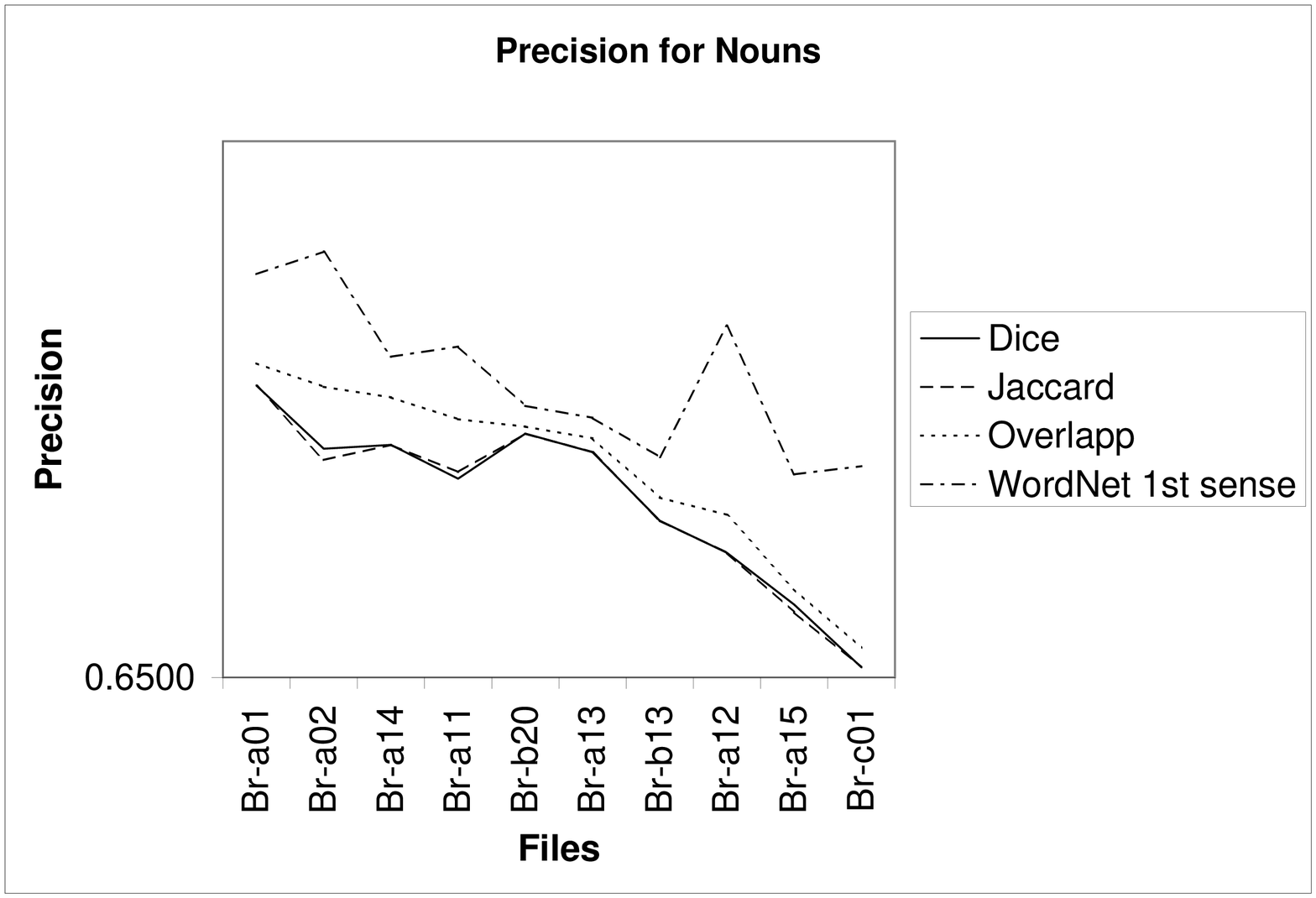}
     \caption{Noun Precision}
    \label{fig:one}
\end{figure}

\begin{footnotesize}
\begin{table}[ht]
    \centering
        \begin{tabular}{|c|c|c|c|c|c|}
            \hline
                \begin{small}{File}\end{small}&\begin{small}{Words}\end{small}&\begin{small}{Dice}\end{small}&\begin{small}{Jaccard}\end{small}&\begin{small}{Overlap}\end{small}&\begin{small}{WN1}\end{small} \\
            \hline
                \begin{small}{Bra01}\end{small}&{486}&{0.758}&{0.758}&{0.767}&{0.800}\\
            \hline
                \begin{small}{Bra02}\end{small}&{479}&{0.735}&{0.731}&{0.758}&{0.808}\\
            \hline
                \begin{small}{Bra14}\end{small}&{401}&{0.736}&{0.736}&{0.754}&{0.769}\\
            \hline
                \begin{small}{Bra11}\end{small}&{413}&{0.724}&{0.726}&{0.746}&{0.773}\\
            \hline
                \begin{small}{Brb20}\end{small}&{394}&{0.740}&{0.740}&{0.743}&{0.751}\\
            \hline
                \begin{small}{Bra13}\end{small}&{399}&{0.734}&{0.734}&{0.739}&{0.746}\\
            \hline
                \begin{small}{Brb13}\end{small}&{467}&{0.708}&{0.708}&{0.717}&{0.732}\\
            \hline
                \begin{small}{Bra12}\end{small}&{433}&{0.696}&{0.696}&{0.710}&{0.781}\\
            \hline
                \begin{small}{Bra15}\end{small}&{354}&{0.677}&{0.674}&{0.682}&{0.725}\\
            \hline
                \begin{small}{Brc01}\end{small}&{434}&{0.653}&{0.653}&{0.661}&{0.728}\\
            \hline
        \end{tabular}
    \caption{Precision for Nouns, sorted descending by
the precision of Overlap measure}
    \label{table:one}
\end{table}
\end{footnotesize}

\begin{figure}[h]
    \centering
        \includegraphics[scale=0.3]{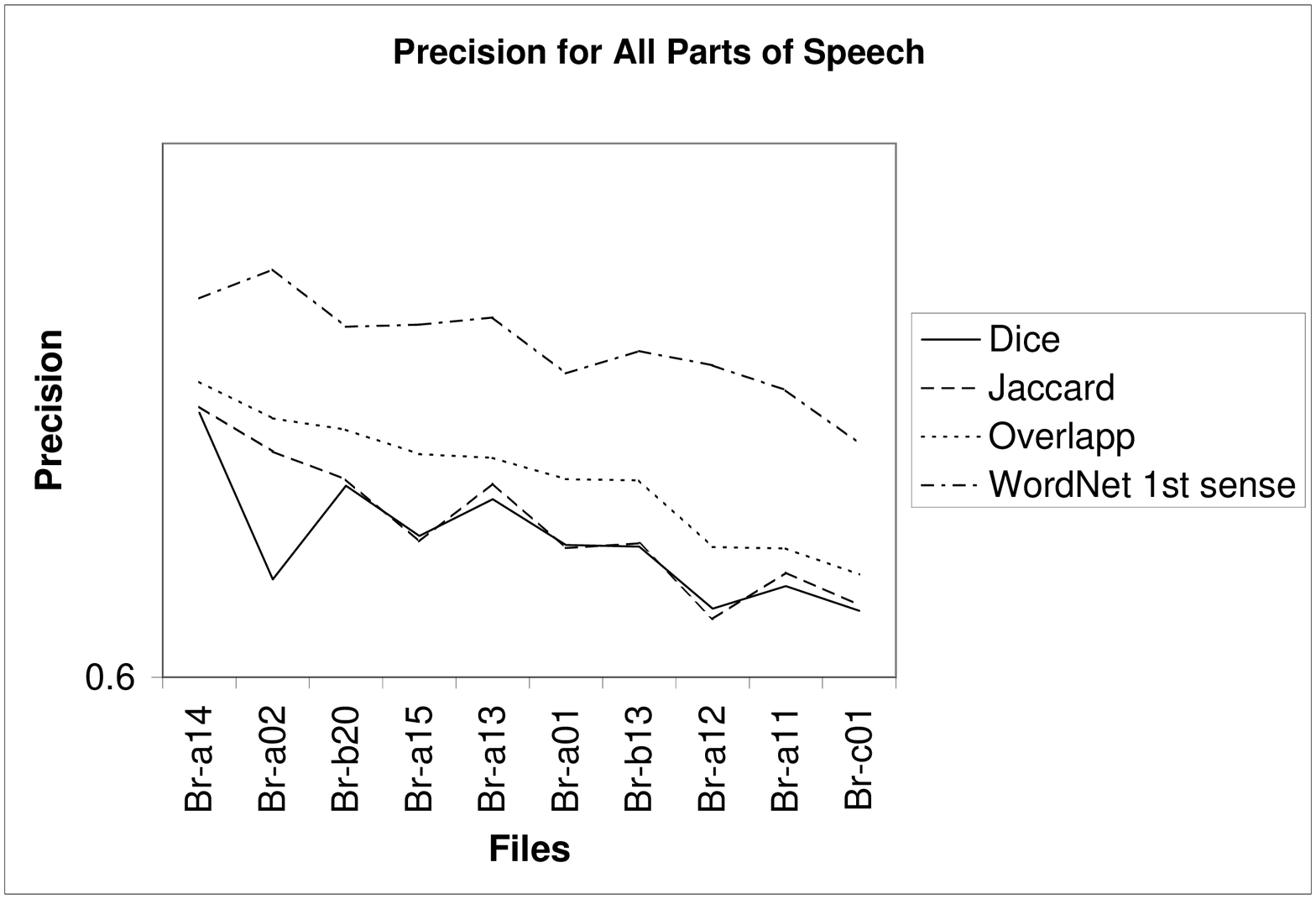}
        \caption{All Parts of Speech Precision}
    \label{fig:two}
\end{figure}
\begin{footnotesize}
\begin{table}[ht]
    \centering
        \begin{tabular}{|c|c|c|c|c|c|}
            \hline
               \begin{small}{File}\end{small}&\begin{small}{Words}\end{small}&\begin{small}{Dice}\end{small}&\begin{small}
               {Jaccard}\end{small}&\begin{small}{Overlap}\end{small}&\begin{small}{WN1}\end{small} \\
            \hline
                \begin{small}{Bra14}\end{small}&{931}&{0.699}&{0.701}&{0.711}&{0.742}\\
            \hline
                \begin{small}{Bra02}\end{small}&{959}&{0.637}&{0.685}&{0.697}&{0.753}\\
            \hline
                \begin{small}{Brb20}\end{small}&{930}&{0.672}&{0.674}&{0.693}&{0.731}\\
            \hline
                \begin{small}{Bra15}\end{small}&{1071}&{0.653}&{0.651}&{0.684}&{0.732}\\
            \hline
                \begin{small}{Bra13}\end{small}&{924}&{0.667}&{0.673}&{0.682}&{0.735}\\
            \hline
                \begin{small}{Bra01}\end{small}&{1033}&{0.650}&{0.648}&{0.674}&{0.714}\\
            \hline
                \begin{small}{Brb13}\end{small}&{947}&{0.649}&{0.650}&{0.674}&{0.722}\\
            \hline
                \begin{small}{Bra12}\end{small}&{1163}&{0.626}&{0.622}&{0.649}&{0.717}\\
            \hline
                \begin{small}{Bra11}\end{small}&{1043}&{0.634}&{0.639}&{0.648}&{0.708}\\
            \hline
                \begin{small}{Brc01}\end{small}&{1100}&{0.625}&{0.627}&{0.638}&{0.688}\\
            \hline
        \end{tabular}
    \caption{Precision for all POS, sorted descending by
the precision of Overlap  measure}
    \label{table:two}
\end{table}
\end{footnotesize}



\begin{figure}[h]
    \centering
        \includegraphics[scale=0.3]{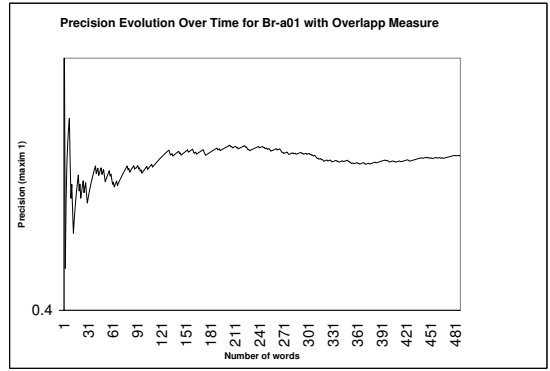}
        \caption{Precision in progress}
    \label{fig:four}
\end{figure}

\section{Applications of CHAD algorithm}

\subsection{Application  to Romanian-English translation}
 \label{app2}

WSD is only an intermediate task in NLP. In Machine Translation
WSD is required for lexical choise  for words that have different
translation for different senses and that are potentially
ambiguous within a given document. However, most  Machine
Translation models do not use explicit WSD \cite{book} (in
Introduction).

    The algorithm implemented by us
    consists in the translation word by word of a Romanian text (using dictionary at
     http://lit.csci.unt.edu/~rada/downloads/RoNLP/R.E. tralexand), then
    the application of chain algorithm to the English text. As the translation of a Romanian
    word in English is multiple, the disambiguation of a triplet is modified as following.
    Let be the word $w_1$ with $k_1$ translations $t_{w_1} ^m$, the word $w_2$ with $k_2$
     translations $t_{w_2} ^n$
and the word $w_3$ with $k_3$ translations $t_{w_3} ^p$. Each
triplet $t_{w_1} ^m t_{w_2} ^n t_{w_3} ^p$ is disambiguated with
the triplet disambiguation algorithm and then the triplet with the
maxim score is selected:
\begin{small}

\hspace{0.2cm}   begin

\hspace{0.3cm}   for   $m=1,k_1$  do

 \hspace{0.5cm} for    $n=1,k_2$ do

 \hspace{0.7cm}        for  $p=1,k_3$ do

\hspace{1cm}        Disambiguate triplet $t_{w_1} ^m t_{w_2} ^n
t_{w_3} ^p$ in  $(t_{w_1} ^m)^* ( t_{w_2} ^n)^* (t_{w_3} ^p)^*$

\hspace{1cm} Calculate  $score ((t_{w_1} ^m)^* ( t_{w_2} ^n)^*
(t_{w_3} ^p)^*)$

\hspace{0.7cm}   endfor

\hspace{0.5cm} endfor

\hspace{0.3cm} endfor

\hspace{0.3cm} Calculate $(m*,n*,p*)=
argmax_{(m,n,p)}score((t_{w_1} ^m)^* ( t_{w_2} ^n)^* (t_{w_3}
^p)^*)$

\hspace{0.3cm} Optimal translation of triplet is $(t_{w_1}
^{m*})^* ( t_{w_2} ^{n*})^*
 (t_{w_3} ^{p*})^*$

\hspace{0.2cm}end

\end{small}

  Let us remark that $(t_{w_1} ^{m*})^*$, for example, is a synset which corresponds to
  the best translation for $w_1$ produced by  CHAD algorithm.
  However, since in Romanian are used many words linked by different spelling signs, these
  composed words are not found in the Romanian-English dictionary. Accordingly, not each
  Romanian word produces an English correspondent as output of the above algorithm. However,
  many translations are  still correct.
         For example, the translation of expression {\it vreme trece} (in the poem "Glossa" of our national
         poet Mihai Eminescu),  is {\it Word: (Rom)vreme
         (Eng)Age$\#n\#4$ , Word: (Rom)trece
         (Eng)$Flow\#v\#1$} . As another example from the same poem,
         where the synset of a word occurs (as an output of our
         application),
          {\it  \c tine toate minte}, is translated in {\it
Word: (Rom) tine   (Eng) $Keep\#v\#8$ :\{keep, maintain\}, Word:
(Rom) toate
         (Eng) $All\#adv\#3$ :\{wholly, entirely, completely, totally, all, altogether,
         whole\},
Word: (Rom) minte
         (Eng) $Judgment\#n\#2$ :\{judgment, judgement, assessment\}}.

\subsection{Application to text entailment verification}

The recognition of text entailment  is one of the most complex
task in Natural Language
   Understanding \cite{tafrestudia}. Thus, a very important problem in some computational
   linguistic applications (as question answering, summarization, segmentation of discourse,
    and others) is to establish if a text
   {\it follows} from  another text.
    For example, a QA system has to identify texts that entail the
    expected answer.
  Similarly, in IR  the concept denoted by a
query expression should be entailed from relevant retrieved
documents. In  summarization, a redundant sentence
 should be entailed from other sentences in the summary.
The application of WSD to text entailment verification is treated
by authors in the paper "Text entailment verification with text
similarity" in this Volume.




\section{Conclusions and further work}

  In this paper we presented a new algorithm of word sense disambiguation.
  The algorithm is parametrized  for:  1. all  words (that means nouns,
   verbs, adjectives, adverbs); 2. all nouns; 3. all verbs. Some experiments
   with this algorithm for ten files of Brown corpus are presented in section 4.2.
The stemming was realized using  the list from
http://snowball.tartarus.org/algorithms/porter/diffs.txt.  The
precision is calculated relative to the corresponding annotated
files in SemCor corpus. Some details of implementation are given
in 4.1.

  We showed in section 5 how   the disambiguation of a text helps in automated
  translation of a text from a language into another
  language: each word in the first text is translated into the most appropriated word
   in the second text. This appropriateness is considered from two points of view:
  1. the point of view of possible translation and 2. the point of view of the real
  sense (disambiguated sense) of the second text. Some experiments with
  Romanian - English translations and text entailment verification
  are given (section 5).

   Another problem which we intend to address in the further work is that of
 optimization of a query in Information Retrieval.
Finding whether a particular sense is connected with an instance
of a word is likely the IR task of finding whether  a document is
relevant to a query. It is established that a good WSD program can
improve performance of retrieval. As IR is used by millions of
users, an average of some percentages of  improvement could be
seen as very significant.

\end{document}